\documentclass[10pt,twocolumn,letterpaper]{article}
\usepackage[pagenumbers]{cvpr}
\usepackage{amsfonts}
\usepackage{amsmath}
\usepackage{amssymb}
\usepackage{booktabs}
\usepackage{caption}
\usepackage{cuted}
\usepackage{diagbox}
\usepackage{epsfig}
\usepackage{float}
\usepackage{graphicx}
\usepackage{microtype}
\usepackage{nicefrac}
\usepackage{pifont}
\usepackage{times}
\usepackage{xcolor}

\usepackage[pagebackref,breaklinks,colorlinks]{hyperref}
\usepackage[capitalize]{cleveref}
\crefname{section}{Sec.}{Secs.}
\Crefname{section}{Section}{Sections}
\Crefname{table}{Table}{Tables}
\crefname{table}{Tab.}{Tabs.}

\usepackage{amsmath,amsfonts,bm}

\def\eqref#1{equation~\ref{#1}}

\def\1{\bm{1}}

\def\eps{{\epsilon}}

\DeclareMathAlphabet{\mathsfit}{\encodingdefault}{\sfdefault}{m}{sl}
\SetMathAlphabet{\mathsfit}{bold}{\encodingdefault}{\sfdefault}{bx}{n}

\newcommand{\R}{\mathbb{R}}

\definecolor{darkred}{rgb}{1,0.5,0.5}

\newcommand{\methodname}{$\textbf{PC}^2$\xspace}
\newcommand{\methodnameplus}{$\textbf{PC}^2\text{-}\textbf{FM}$\xspace}
\newcommand{\methodnameplusminus}{$\textbf{PC}^2\text{-}\textbf{FA}$\xspace}
\newcommand{\colorhref}[3][blue]{\href{#2}{\color{#1}{#3}}}%

\makeatletter
\renewcommand{\paragraph}{%
\@startsection{paragraph}{4}%
{\z@}{0.25em}{-1em}%
{\normalfont\normalsize\bfseries}%
}
\makeatother

\def\cvprPaperID{9217}
\def\confName{CVPR}
\def\confYear{2023}

\title{\methodname: Projection-Conditioned Point Cloud \\ Diffusion for Single-Image 3D Reconstruction}

\author{Luke Melas-Kyriazi
\quad Christian Rupprecht
\quad Andrea Vedaldi
\\
\\
\small Visual Geometry Group, Department of Engineering Science, University of Oxford\\
\small {\texttt{\{lukemk,chrisr,vedaldi\}@robots.ox.ac.uk}}\\
\small {\colorhref{https://lukemelas.github.io/projection-conditioned-point-cloud-diffusion}{https://lukemelas.github.io/projection-conditioned-point-cloud-diffusion}}
}

\begin{document}
\maketitle

\begin{strip}
\vspace*{-2mm}
\centering
\includegraphics[width=0.95\linewidth]{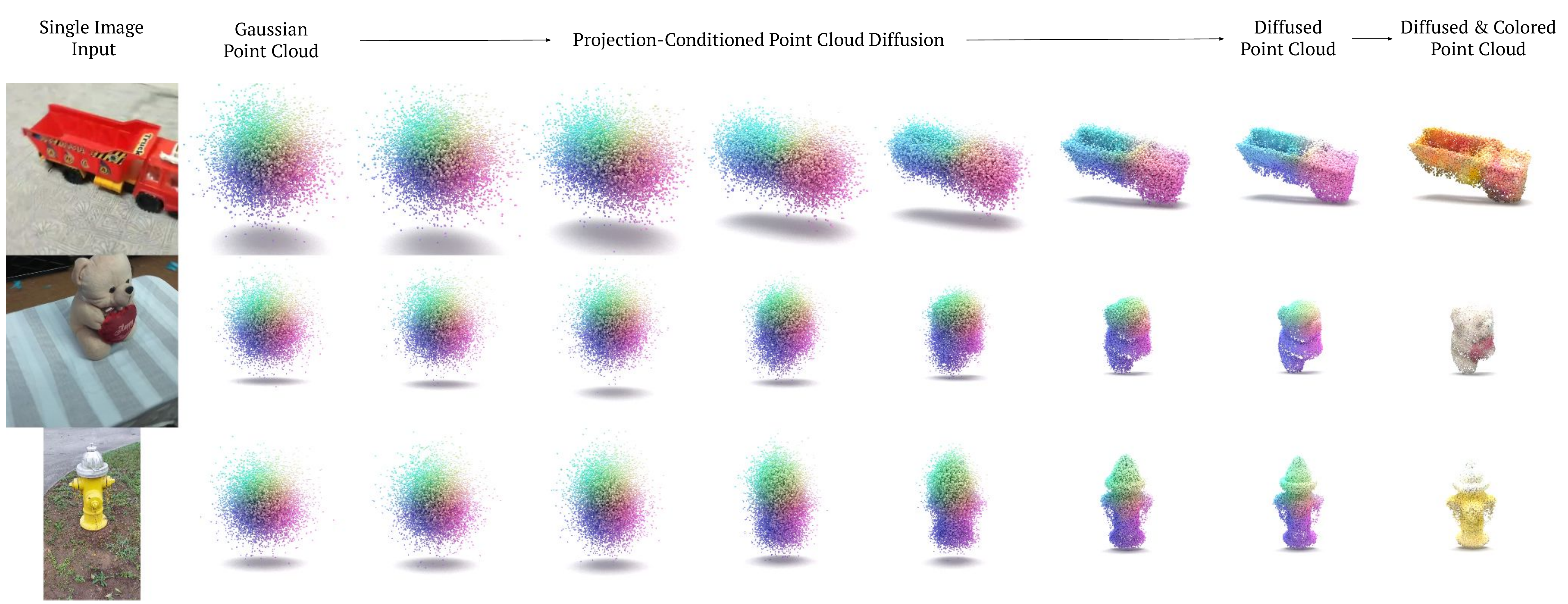}
\captionof{figure}{\textbf{Projection-Conditioned Point Cloud
Diffusion:} Our model performs single-image 3D point cloud reconstruction by gradually diffusing an initially random point cloud to align the with input image. Our model has been trained through simple, sparse COLMAP supervision from videos.}
\label{f:splash}
\end{strip}

\begin{abstract}
Reconstructing the 3D shape of an object from a single RGB image is a long-standing and highly challenging problem in computer vision.
In this paper, we propose a novel method for single-image 3D reconstruction which generates a sparse point cloud via a conditional denoising diffusion process.
Our method takes as input a single RGB image along with its camera pose and gradually denoises a set of 3D points, whose positions are initially sampled randomly from a three-dimensional Gaussian distribution, into the shape of an object.
The key to our method is a geometrically-consistent conditioning process which we call \textit{projection conditioning}:
at each step in the diffusion process, we project local image features onto the partially-denoised point cloud from the given camera pose.
This projection conditioning process enables us to generate high-resolution sparse geometries that are well-aligned with the input image, and can additionally be used to predict point colors after shape reconstruction.
Moreover, due to the probabilistic nature of the diffusion process, our method is naturally capable of generating multiple different shapes consistent with a single input image.
In contrast to prior work, our approach not only performs well on synthetic benchmarks, but also gives large qualitative improvements on complex real-world data.
\end{abstract}
\section{Introduction}\label{s:intro}

Reconstructing the 3D structure of an object from a single 2D view is a long-standing computer vision problem. 
Given more than one view, it is possible to reconstruct an object's shape using the tools of multiple-view geometry, but in the case of a single view the problem is highly ill-posed and requires prior understanding of the possible shapes and appearances of the objects.

Despite the difficulty of this task, humans are adept at using a range of monocular cues and prior knowledge to infer the 3D structure of common objects from single views.
For example, when a person looks at a photograph of an object, they can easily imagine what the shape of the backside of the object may plausibly be.
The ease with which humans perform this task, and its relative difficulty for computer vision systems, has attracted the attention of the research community for nearly half a century~\cite{kruppa1913investigation,maybank85the-angular,hartley00multiple,hartley04reconstruction}.
Moreover, given the prevalence of single-view RGB image data in the real world, the single-view reconstruction problem has practical applications in numerous areas, such as augmented and virtual reality.

Over the past five years, a substantial body of research has emerged around reconstruction via end-to-end deep learning methods~\cite{sun2018pix3d,tatarchenko2019single,wallace2019few,wang2018pixel2mesh,mescheder2019occupancy,yagubbayli2021legoformer,jang2021codenerf,rematas2021sharf,liu2019soft,yu2021pixelnerf,xia2019realpoint,kato2019learning,Lin_2019_CVPR,Yu_2019_CVPR,xie2020pix2vox++,xie2019pix2vox}.
These works are capable of predicting volumes from single images, yet many remain limited to low-resolution geometries (\eg dense voxel grids), and thus have limited ability to reconstruct the precise geometry of the input image.
Some recent works~\cite{yu21pixelnerf:,jang21codenerf:,rematas21sharf:,Henzler_2021_CVPR} also utilize implicit representations and radiance fields, which are capable of rendering novel views with photographic quality but often suffer from other drawbacks, such as an inability to reconstruct a distribution of possible 3D shapes from a single input image.

In this work, we take inspiration from recent progress in the generative modeling of 2D images using denoising diffusion probabilistic models.
In the domain of 2D images, we have seen that diffusion models (\eg Latent-Diffusion~\cite{rombach2021highresolution}, SDEdit~\cite{meng2022sdedit}, GLIDE~\cite{nichol2021glide}, and DALL-E 2~\cite{ramesh2022hierarchical}) can produce remarkably high-fidelity image samples either from scratch or when conditioned on textual inputs.
They do so by learning complex priors over the appearances of common objects.
For example, GLIDE~\cite{nichol2021glide} is designed to conditionally inpaint unseen regions in an image.
We seek to bring these advances to the domain of 3D reconstruction to conditionally generate the shape of unseen regions of a 3D object.

In order to do so, we propose to use diffusion models for single-image 3D reconstruction.
Our approach represents shapes as unstructured point clouds:
we gradually denoise a randomly-sampled set of points into a target shape conditional on an input image and its corresponding viewpoint.
The key to our approach is a novel way of conditioning the diffusion process to produce 3D shapes which are geometrically consistent with the input images.
Specifically, at each iteration of the diffusion process, we project image features directly onto the points of the partially-denoised point cloud.
When shape reconstruction is complete, we are also able to use the same projection conditioning to predict point colors.

Our method differs from prior work in three major aspects:
(1) our point cloud-based shape representation,
(2) our projection-conditioned diffusion model, and
(3) our use of projection conditioning to predict point colors in addition to point shapes.
Although these may seem separate from one another, they are in fact directly linked: the order-agnostic and unstructured nature of point clouds naturally lends itself to the highly flexible nature of diffusion models.
Without the point cloud-based representation, our projection-conditioned diffusion approach would be limited to coarse voxel-based shape approximation.

Finally, due to its probabilistic nature, given a single input image, our model is able to generate multiple plausible 3D point clouds which are all consistent with the input.
We leverage this property by introducing a novel filtering step in the sampling process, which helps to address the ill-posed nature of the single-view 3D reconstruction problem.
Specifically, we generate multiple point clouds for a given input image, and filter these point clouds according to how well they match the input mask.
Filtering enables us to benefit from the diversity of our model's reconstructions in an entirely automated fashion.

Experimentally, we not only perform competitively on the synthetic ShapeNet benchmark, but we also move beyond synthetic data: we demonstrate high-quality qualitative single-view reconstruction results on multiple categories in the challenging, real-world Co3D~\cite{reizenstein21co3d} dataset.

We present this work as a first step toward using denoising diffusion models for large-scale 3D reconstruction.
Given the success of scaling 2D diffusion models over the past two years, our work demonstrates that 3D diffusion models could represent a path toward large-scale 3D reconstruction of complex objects from single-view images.
To encourage further research into diffusion models for 3D reconstruction, we will release code and pretrained models on acceptance.

\begin{figure*}
\centering
\includegraphics[width=0.94\textwidth]{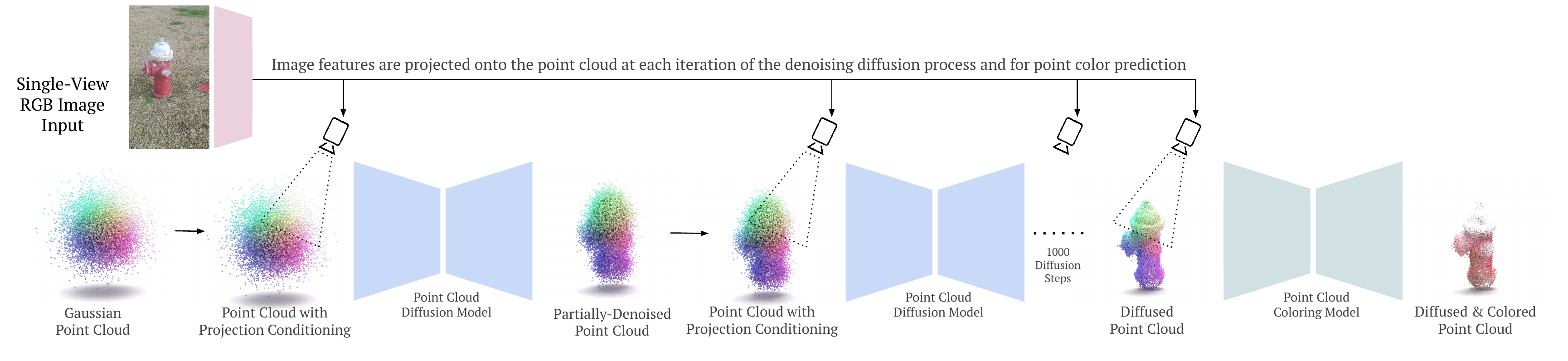}
\caption{
\methodname reconstructs a colored point cloud from a single input image along with its camera pose.
The method contains two sub-parts, both of which utilize our model projection conditioning method.
First, we gradually denoise a set of points into the shape of an object.
At each step in the diffusion process, we project image features onto the partially-denoised point cloud from the given camera pose, augmenting each point with a set of neural features.
This step makes the diffusion process conditional on the image in a geometrically-consistent manner, enabling high-quality shape reconstruction.
Second, we predict the color of each point using a model based on the same projection procedure.
}%
\label{fig:method}
\end{figure*}
\section{Related work}\label{s:related}

\paragraph{Single-View 3D reconstruction}

Originally, research on 3D reconstruction focused primarily on reconstruction from multiple views using classical geometric techniques~\cite{Hartley2004}.
Most commonly, these methods map the input image into a feature representation using a 2D convolutional network, and then decode the features into a 3D representation using a 3D convolutional network or a sequence model.
This decoding process generally requires the output 3D representation to have a fixed and regular structure, such as a voxel grid.

One of the pioneering methods in this line of work was 3D-R2N2 \cite{choy20163d}, which encodes an input 2D image into a low dimensional embedding using a standard 2D convolutional network, process this embedding with a 3D-LSTM, and then decode an output voxel grid with a 3D convolutional network. \cite{choy20163d} also introduced a standard set of views and renders for evaluating single-view reconstruction methods based on the ShapeNet dataset, which we use for evaluation in this work (\cref{s:experiments}). LSM \cite{kar2017learning} processes images into feature maps with a 2D network, unprojects these features into a 3D voxel grid, and processes the grid with a 3D convolutional network. Pix2Vox \cite{xie2019pix2vox} and its successor Pix2Vox++ \cite{xie2020pix2vox++} use a simple encoder-decoder architecture consisting of a 2D convolutional encoder and a 3D convolutional decoder augmented with a multi-scale fusion module. Most recently, LegoFormer~\cite{yagubbayli2021legoformer} adopts a transformer-based approach. They encode an image into a feature vector and decode a $32^3$ dense voxel grid as a sequence of 512 $4 \times 4$ blocks. For this decoding they employ a non-autoregressive transformer-based decoder with learned queries. 

Additionally, over the past two years a new body of research has emerged around differentiable rendering.
Popularized by NeRF~\cite{mildenhall2020nerf}, these methods learn implicit representations of either radiance~\cite{mildenhall2020nerf} or signed distance~\cite{wang21neus:}.
Although the vast majority of this work operates on the level of single scenes with abundant multi-view data, a few recent works have tackled the single-view setting~\cite{Henzler_2021_CVPR,yu2021pixelnerf,jang21codenerf:,rematas21sharf:,kulhanek22viewformer:}.
Most relevant for this paper are Nerf-WCE~\cite{Henzler_2021_CVPR} and PixelNeRF~\cite{yu2021pixelnerf}, which concurrently developed methods for single/few-view reconstruction, conditioning a NeRF on image features from reference views. Both work well in the few-view settings on datasets such as Co3D~\cite{reizenstein21common}, but strong results in the single-view setting remain elusive. 
We compare qualitatively against NeRF-WCE in~\cref{s:experiments}.

In this paper, we take an entirely different approach from those above.
We utilize the flexible machinery of denoising diffusion probabilistic models for point cloud-based reconstruction.
Due to their probabilistic nature, diffusion models enable us to capture the ambiguity of unseen regions while also generating high-resolution point cloud-based shapes.
Below, we give an overview of the related literature on diffusion models.

\paragraph{Diffusion Models}

Denoising diffusion probabilistic models are a class of generative models based on iteratively reversing a Markovian noising process.
Early work on diffusion models for image generation, inspired by thermodynamics, framed the problem as one of optimizing a variational lower bound of a latent variable model~\cite{ddpm}.
Later work showed that this modeling paradigm is equivalent to score-based generative modeling~\cite{song2019generative,song2020improved}, and can also be seen as a discretization of a continuous process based on stochastic differential equations~\cite{song2021scorebased}.
Recent work has explored faster/deterministic sampling for diffusion models~\cite{liu2021pseudo,watson2021learning,ddim}, class-conditional models~\cite{dhariwal2021diffusion,song2020improved}, text-conditional models~\cite{nichol2021glide}, and modeling in latent space~\cite{rombach2021highresolution}.

\paragraph{Diffusion Models for Point Clouds}

Over this past year, three applications of diffusion models for unconditional point cloud generation have emerged.
{}\cite{luo2021diffusion} and~\cite{Zhou_2021_ICCV} proposed similar generation setups, differing in their use of a PointNet~\cite{qi2017pointnet} and a Point-Voxel-CNN~\cite{Zhou_2021_ICCV}, respectively.
Most recently~\cite{lyu2022a} uses a diffusion model followed by a refinement model for the purpose of completing partial point clouds.
However, these three works only tackle the problem of unconditional shape generation or completion and do not address the question of how to reconstruct real-world images.
Additionally, they train only on synthetic datasets, whereas we show results for complex real-world data.

\begin{figure*}[ht]
\centering
\includegraphics[width=\textwidth]{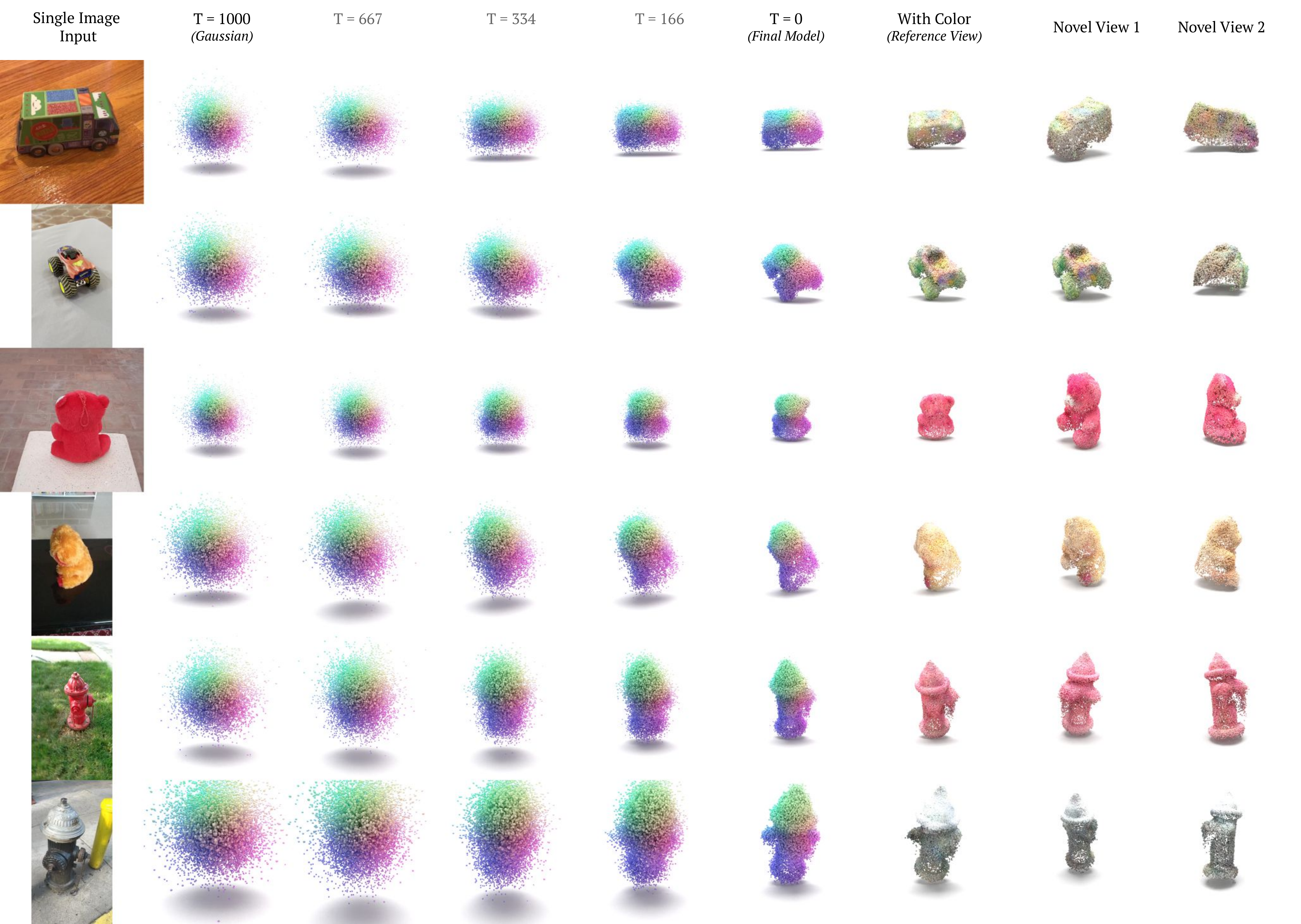}
\caption{Examples of \methodname on real-world images from Co3D~\cite{reizenstein21co3d}, along with intermediate steps in the diffusion process.
The first column shows input images, while the subsequent five columns show the evolution of the point cloud from a randomly sampled Gaussian to a final shape over the course of the diffusion process.
The following column shows the result of our coloring model applied to our reconstructed shape.
The final two columns show renders of our colored point cloud from two randomly sampled novel views, demonstrating that it is a complete 3D shape.
Our method is capable of reconstructing objects from real-world images with challenging viewpoints.
}%
\label{fig:qualitative}
\end{figure*}

\begin{table}[t]
\small
\centering
\begin{tabular}{lccc}
\toprule
\textit{Category} &  Global Cond. & \textbf{\methodname} \\ \midrule
airplane    & 0.197 & \textbf{0.473} \\
table       & 0.156 & \textbf{0.290} \\
\bottomrule
\end{tabular}
\caption{\textbf{Motivating Experiment.} Here, we show a small quantitative comparison on two categories from ShapeNet~\cite{choy20163d} which motivates our projection conditioning method. The table shows F-scores (higher is better). For the global conditioning approach, the input image is first processed into a global feature vector using a convolutional network~\cite{simonyan2014very}, and this vector is used to condition a denoising diffusion model. Relative to the global conditioning, our projection conditioning is much more effectively able to reconstruct object geometries.
}%
\label{tab:comparison_global}
\end{table}

\section{Method}\label{s:method}

In the following sections we describe our method.
We begin with an overview of denoising diffusion models, which forms the foundation of our method.
Next, we introduce our novel conditioning scheme \methodname.
Finally, we describe our filtering method \methodnameplus, which utilizes the probabilistic nature of diffusion to address the ill-posed nature of the single-view 3D reconstruction problem.

\subsection{Diffusion Models}

Diffusion denoising probabilistic models are general-purpose generative models inspired by stochastic differential equations and non-equilibrium thermodynamics.
Diffusion denoising models are based on an iterative noising process, which adds noise to a sample $X_0 \sim q(X_0)$ from a target data distribution $q(X_0)$ over a series of steps.
The stepsize of each step of this process is given by a variance schedule $\{\beta_t\}_{t=0}^{T}$:
\[
q(X_t|X_{t-1}) = \mathcal{N}(X_t; \sqrt{1-\beta_t}X_{t-1}, \beta_t \mathbf{I})
\]
Each $q(X_t|X_{t-1})$ is a Gaussian distribution, implemented by using the the well-known reparameterization trick:
\[
q(X_t|X_0) = \sqrt{\bar{\alpha}_t} X_0 + \epsilon \sqrt{1-\bar{\alpha}_t}
\]
where $\alpha_t = 1 - \beta_t$, $\bar{\alpha}_t = \prod_{s=0}^{t} \alpha_s$, and $\epsilon \sim \mathcal{N}(0, \mathbf{I})$.

In order to form a generative model, we consider the \textit{reverse} diffusion process, which begins with a sample from the noise distribution $q(X_T)$ and then denoises over a series of steps to obtain a sample from the target distribution $q(X_0)$.
Confusingly, this reverse process is sometimes also referred to as the ``diffusion process.''
In order to reverse the diffusion process, we learn the distribution $q(X_{t-1}|X_t)$ using a neural network $s_{\theta}(X_{t-1}|X_t) \approx q(X_{t-1}|X_t)$.
We can then (approximately) sample from $q(X_0)$ by starting with a sample $X_T \sim q(X_T)$ and then iteratively sampling from $q(X_{t-1}|X_t)$.
When the sampling step size is sufficiently small, $q(X_{t-1}|X_t)$ is well approximated by an isotropic Gaussian with a fixed small covariance.
Thus, one only has to predict the mean of $q(X_{t-1}|X_t)$, which in practice is parameterized as predicting the noise $\eps$ with a neural network $s_{\theta}(X_{t-1}|X_t)$.

\subsection{Point Cloud Diffusion Models}

We consider 3D point cloud with $N$ points as a $3N$ dimensional object and learn a diffusion model $s_{\theta}: \R^{3N} \to \R^{3N}$. 
This network denoises the positions of a set of points from a spherical Gaussian ball into a recognizable object. At a very high level, at each step, we predict the offset of each point from its current location, and we iterate this process to arrive at a sample from our target distribution $q(X_0)$.

Concretely, the network is trained to predict the noise $\eps \in \R^{3N}$ added in the most recent time step using an $L_2$ loss between the true and predicted noise values:
\[
\mathcal{L} = E_{\epsilon \sim \mathcal{N}(0, \mathbf{I})}\left[\|\epsilon - s_{\theta}(X_t, t)\|_2^2\right]
\]

At inference time, a random point cloud $X_T \sim N(\textbf{0}, \textbf{I}_{3N})$ is sampled from a $3N$-dimensional Gaussian and the reverse diffusion process is run to produce a sample $X_0$.
At each step, we recover the mean $\mu_{\theta,t}$ of our approximation of $q(X_{t-1}|X_t)$ from the prediction $s_{\theta}(X_t, t)$, and we use this mean to sample from $q(X_{t-1}|X_t)$.

\subsection{Conditional Point Cloud Diffusion Models}

We formulate 3D reconstruction as conditional generation: the target distribution is the conditional distribution $q(X_0 | I, V)$ where $I$ is an input image and $V$ is the corresponding camera view.
The key question of this paper is how exactly one should condition the model on the reference image and camera view. 
The most immediate way of conditioning would be to 
provide a global learned embedding of the input image as an auxilliary input to the denoising function $s_\theta$.
This setup resembles prior work which used encoder-decoder architectures conditioned on image embeddings to generate 3D shapes~\cite{choy20163d,xie2019pix2vox,xie2020pix2vox++}.

However, this approach is lacking in that it only promotes a weak form of geometric consistency between the input image and the reconstructed shape.
Empirically, preliminary experiments showed that it often generated plausible shapes, but that these shapes did not always match the input image from the given view (see \cref{tab:comparison_global}).

This finding matches up precisely with the findings of~\cite{Tatarchenko_2019_CVPR}, who analyzed voxel-based methods which decode from a global feature vector in a similar manner: they found that these networks primarily performed a sort of coarse classification of the input image into a small number of shape categories and then generated a generic object from that category.
Although coarse classification may produce adequate results in a highly curated synthetic setting such as ShapeNet, it will inevitably fail to capture the complexity of objects in real-world scenes.
These insights led us to develop \methodname, a geometrically consistent method for locally-conditional point cloud diffusion.

\subsection{\methodname: Projection-Conditional Diffusion Models}

To address the geometric consistency issues observed above, our method \methodname instead \textit{projects} the image onto the partially-denoised point at each step in the diffusion process.
We first process the image into a dense feature volume using a standard 2D image model(\eg a CNN or ViT~\cite{dosovitskiy2020image}). Next, before each diffusion step, we project these features onto the point cloud, attaching a neural feature to each point. Crucially, each point is augmented with a \textit{different} neural feature, as determined by its location in space relative to the input camera and the input image. 

Formally, let $I \in \R^{H\times W \times C}$ be the feature volume produced by our 2D image model applied to our input image, where $C$ is the number of feature channels. $s_\theta$ is now a function $\R^{(3+C) N} \to \R^{3 N}$ which predicts the noise $\eps$ from the augmented point cloud $X_t^{+} = [X_t, X_t^{proj}]$.
The projected features $X_t^{proj}$ are given by
$
X_t^{proj} = \mathcal{P}_{V_I}(I, X_t)
$
where $\mathcal{P}_{V_I}$ is the projection function from camera view $V_I$, $I$ is the input image, and $X_t$ is the partially-noised point cloud.

One straight-forward approach to design $\mathcal{P}_{V_I}(I, X_t)$ would be to simply project the 3D points on the image and take the corresponding image features.
However, this is akin to assuming that the point cloud is transparent.
Instead, we found it beneficial to choose a projection function that properly accounts for self-occlusion.
We implement it by rasterizing the points, assuming that they have some small, non-zero radius $\rho$.
That is, $\mathcal{P}_{V_I}$ is a rasterization procedure~\cite{ravi2020accelerating}.
Due to highly optimized rasterization routines in modern software, this process is very efficient; it accounts for a negligible fraction of training time and computation. Full details about the rasterization process and the projected information are provided in the supplementary material.

Finally, we find that we can apply \textit{exactly the same} projection-based conditioning to reconstruct object color in addition to object shape.\footnote{
Learning a coloring network is beneficial compared to a simple projection of colors from the image to points because a coloring network could learn to color an object backside differently from its front size (for example, in the case of a teddy bear).
}
Specifically, we learn a separate coloring network $c_{\theta}: \R^{(3+C) N} \to \R^{C N}$, which takes as input a point cloud augmented with projection conditioning, and outputs the colors of each point.
Empirically, we found that a single-step coloring model produced results comparable to a diffusion model. To reduce computational complexity, we hence utilize the single step model. :

\begin{table}[t]
\centering
\makebox[\linewidth]{

\begin{tabular}{lccccc}
\toprule
\textit{Category}          & {}\cite{choy20163d} & {}\cite{yagubbayli2021legoformer} & {}\cite{xie2020pix2vox++} & \textbf{\methodname} & \textbf{\methodnameplus} \\ \midrule
airplane         &  0.225 &  0.215 &  0.266 &  0.473 &  \textbf{0.589} \\
bench            &  0.198 &  0.241 &  0.266 &  0.305 &  \textbf{0.334} \\
cabinet          &  0.256 &  0.308 &  \textbf{0.317} &  0.203 &  0.211 \\
car              &  0.211 &  0.220 &  0.268 &  0.359 &  \textbf{0.372} \\
chair            &  0.194 &  0.217 &  0.246 &  0.290 &  \textbf{0.309} \\
display          &  0.196 &  0.261 &  \textbf{0.279} &  0.232 &  0.268 \\
lamp             &  0.186 &  0.220 &  0.242 &  0.300 &  \textbf{0.326} \\
loudspeaker      &  0.229 &  0.286 &  \textbf{0.297} &  0.204 &  0.210 \\
rifle            &  0.356 &  0.364 &  0.410 &  0.522 &  \textbf{0.585} \\
sofa             &  0.208 &  0.260 &  \textbf{0.277} &  0.205 &  0.224 \\
table            &  0.263 &  0.305 &  \textbf{0.327} &  0.270 &  0.297 \\
telephone        &  0.407 &  0.575 &  \textbf{0.582} &  0.331 &  0.389 \\
watercraft       &  0.240 &  0.283 &  0.316 &  0.324 &  \textbf{0.341} \\ \midrule
\textit{Average} &  0.244 &  0.289 &  0.315 &  0.309 &  \textbf{0.343} \\
\bottomrule
\end{tabular}

}
\caption{
Comparison of single-view 3D reconstruction performance with prior work on ShapeNet-R2N2.
F-Score is calculated using a distance threshold of $0.01$~\cite{xie2019pix2vox}.
Our method performs similarly to prior work without filtering and outperforms prior work using filtering (both \methodnameplusminus and \methodnameplus);
this filtering step leverages the method's ability to produce multiple reconstructions for a given input image.
}%
\label{tab:comparison_others}
\end{table}

\begin{table}[t!]
\centering
\vspace{5mm}
\begin{tabular}{lcccc}
\toprule
\textit{Category} &  \textbf{\methodname} &  \textbf{\methodnameplusminus} &  \textbf{\methodnameplus} &  \textbf{Oracle} \\ \midrule
airplane             & 0.473 &     0.517          & 0.589 &   0.681 \\
bench                & 0.305 &     0.316          & 0.334 &   0.444 \\
cabinet              & 0.203 &     0.246          & 0.211 &   0.303 \\
car                  & 0.359 &     0.369          & 0.372 &   0.420 \\
chair                & 0.290 &     0.312          & 0.309 &   0.377 \\
display              & 0.232 &     0.277          & 0.268 &   0.357 \\
lamp                 & 0.300 &     0.320          & 0.326 &   0.399 \\
loudspeaker          & 0.204 &     0.235          & 0.210 &   0.288 \\
rifle                & 0.522 &     0.538          & 0.585 &   0.686 \\
sofa                 & 0.205 &     0.242          & 0.224 &   0.298 \\
table                & 0.270 &     0.293          & 0.297 &   0.420 \\
telephone            & 0.331 &     0.400          & 0.389 &   0.523 \\
watercraft           & 0.324 &     0.322          & 0.341 &   0.424 \\ \midrule
\textit{Average}     & 0.309 &     0.337          & 0.343 &   0.432 \\
\bottomrule
\end{tabular}
\caption{Single-view 3D reconstruction performance on ShapeNet-R2N2 for different levels of filtering.
\methodname produces a single sample for each input image.
\methodnameplus (see \cref{s:method}) uses $5$ samples for automated $IoU$-based filtering.
The ``Oracle'' method evaluates the upper bound of filtering: it computes the $F$-score of $5$ samples using the ground truth point cloud and chooses the best sample for each input image.
Filtering, both without object masks (\methodnameplusminus) and with masks (\methodnameplus) improves results substantially, but does not fully close the gap to the oracle.}%
\label{tab:comparison_oracle}
\end{table}

\begin{table}[t!]
\centering
\begin{tabular}{lccccc}
\toprule
\textit{Samples}  &     1 &     2 &     3 &     4 &     5 \\ \midrule
\textit{Avg. F-Score} & 0.310 & 0.330 & 0.335 & 0.340 & \textbf{0.343} \\
\bottomrule
\end{tabular}
\caption{
Performance of \methodnameplus on ShapeNet-R2N2 when filtering different numbers of samples.
There is a large improvement (0.02 $F$-score) from a single sample (\ie, no filtering) to two samples. 
Performance continues to increase with diminishing returns.}%
\label{tab:comparison_filtering}
\vspace*{-3mm}
\end{table}

\subsection{\methodnameplus and $\textbf{PC}^2\text{-}\textbf{FA}$: Filtering for 3D Reconstruction}

Due to the ill-posed nature of the single-view 3D reconstruction problem, there are many possible ground truth shapes corresponding to each input image.
Differently from most prior works~\cite{yu21pixelnerf:,xie2019pix2vox,xie2020pix2vox++,yagubbayli2021legoformer}, our model is inherently probabilist, and thus admits sampling multiple different reconstructions from a single input. 

We take advantage of this property by proposing to generate \textit{multiple independent samples} for each input image and then filter these according to an automated criterion. Thus, we are able to generate a diverse set of outputs and choose the one which is ``most plausible'' for a given input image.

In practice, we propose two simple criteria for filtering, both of which involve the object silhouette. One of these uses additional mask supervision, while the other does not. In both cases, we begin by rendering each point cloud $\{X_0^{(i)}\}_{i=1}^{N}$ from the input camera view $V$. We render the points as balls with a small fixed radius, such that we obtain the silhouette $\hat{M}^{(i)}$ of our point cloud sample. 

\paragraph*{With mask supervision (\methodnameplus).} We compare each silhouette $\hat{M}$ with the object mask $M$, which is extracted automatically using Mask-RCNN~\cite{he17mask}. We then calculate their intersection-over-union $IoU(\hat{M}^{(i)}, M)$ and select the sample with the highest $IoU$ score. 

\paragraph*{Without mask supervision (\methodnameplusminus).} Rather than using an object mask, we filter based on the \textit{mutual agreement} between our predictions. Specifically, we compute the IoU of each prediction with all of \textit{the other predictions} and select the mask with the highest average $IoU$. In this way, we are able to select high-quality masks without any additional supervision. 

Finally, we emphasize that our filtering approach is a general one; the two methods explored here are by no means comprehensive.

\begin{figure}
\centering
\includegraphics[width=\linewidth]{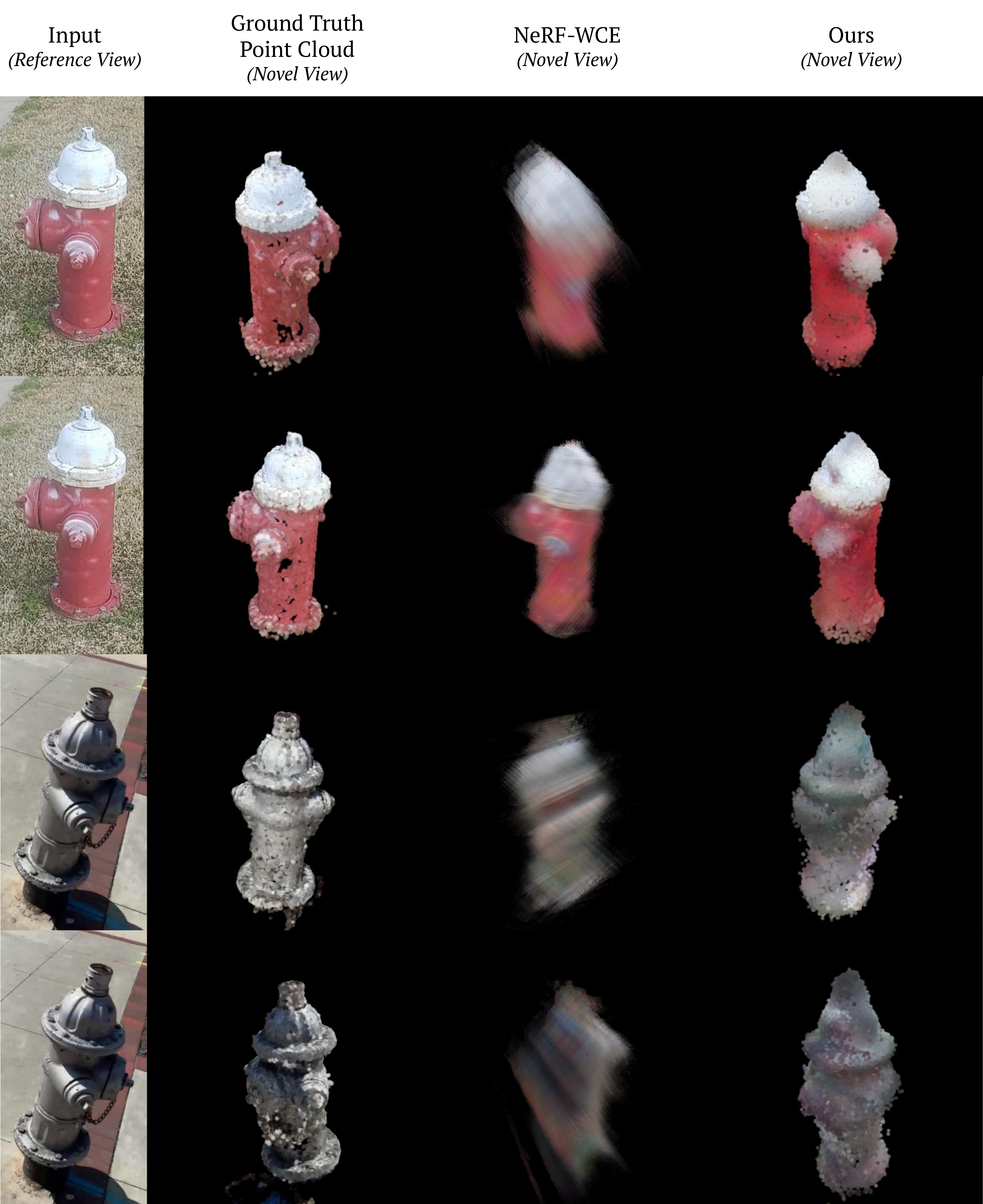}
\caption{
Comparison with NeRF-WCE~\cite{Henzler_2021_CVPR} on the hydrants category of Co3D~\cite{reizenstein21co3d}.
The first column shows the reference image, while the following three images show renders from a novel view (sampled randomly from a circle).
The second column shows a render of the ground truth point cloud (obtained from COLMAP on the entire video).
The third and fourth columns show renders from NeRF-WCE and our model.
NeRF-WCE produces highly blurry outputs due to its deterministic nature.
By contrast, our model produces realistic shapes  from any viewpoint.}%
\label{fig:qualitative_nerf}
\end{figure}
\begin{figure}
\centering
\includegraphics[width=\linewidth]{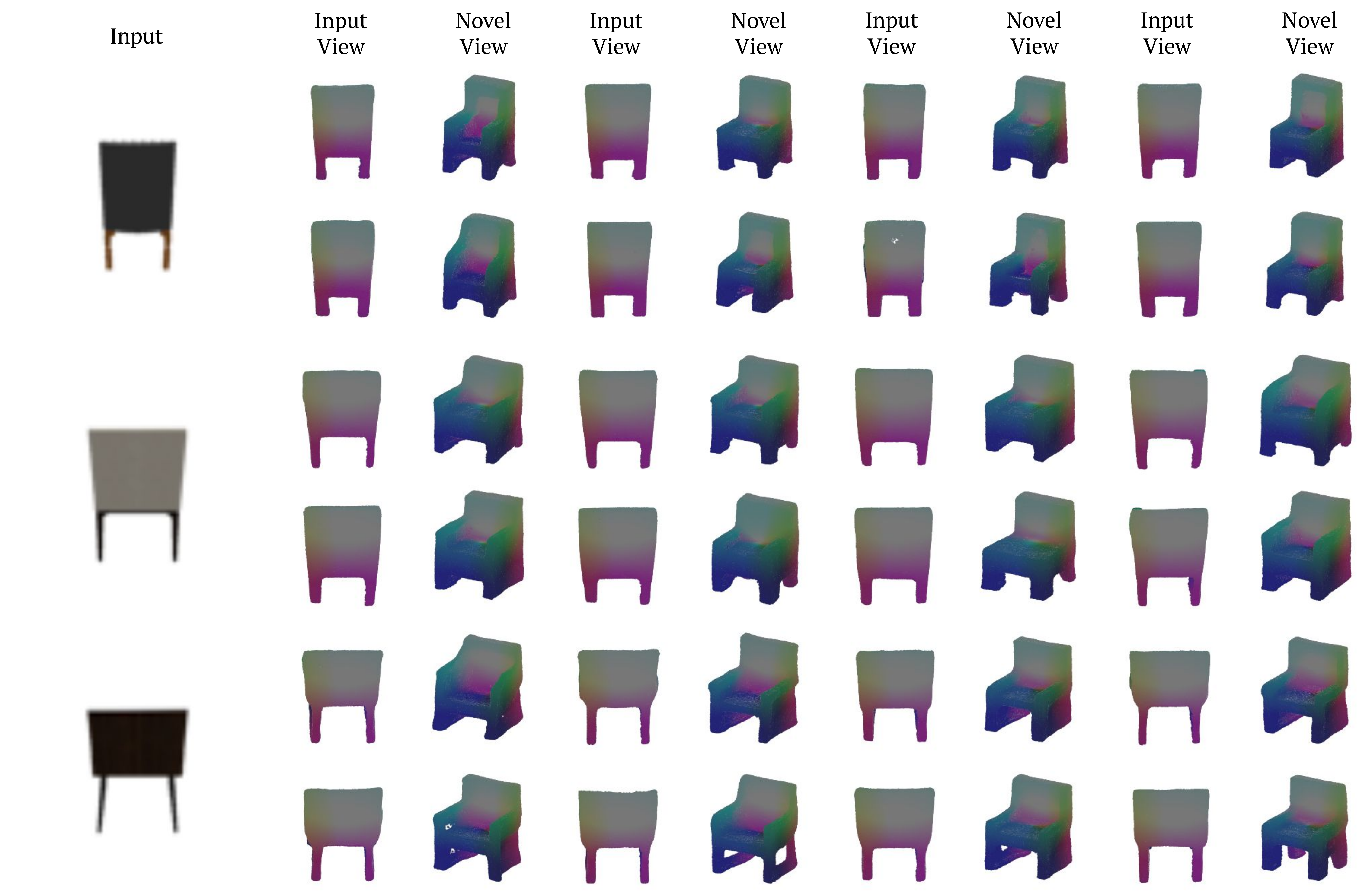}
\caption{Diversity.
The leftmost column shows a reference image from the chair category of ShapeNet, chosen specifically for its high shape ambiguity.
The following images show generations produced by our model for eight random seeds.
We see that our model is capable of producing significant variations in shape while still matching the input image from the reference viewpoint.}%
\label{fig:diversity}
\end{figure}
\begin{figure}
\centering
\includegraphics[width=0.45\textwidth]{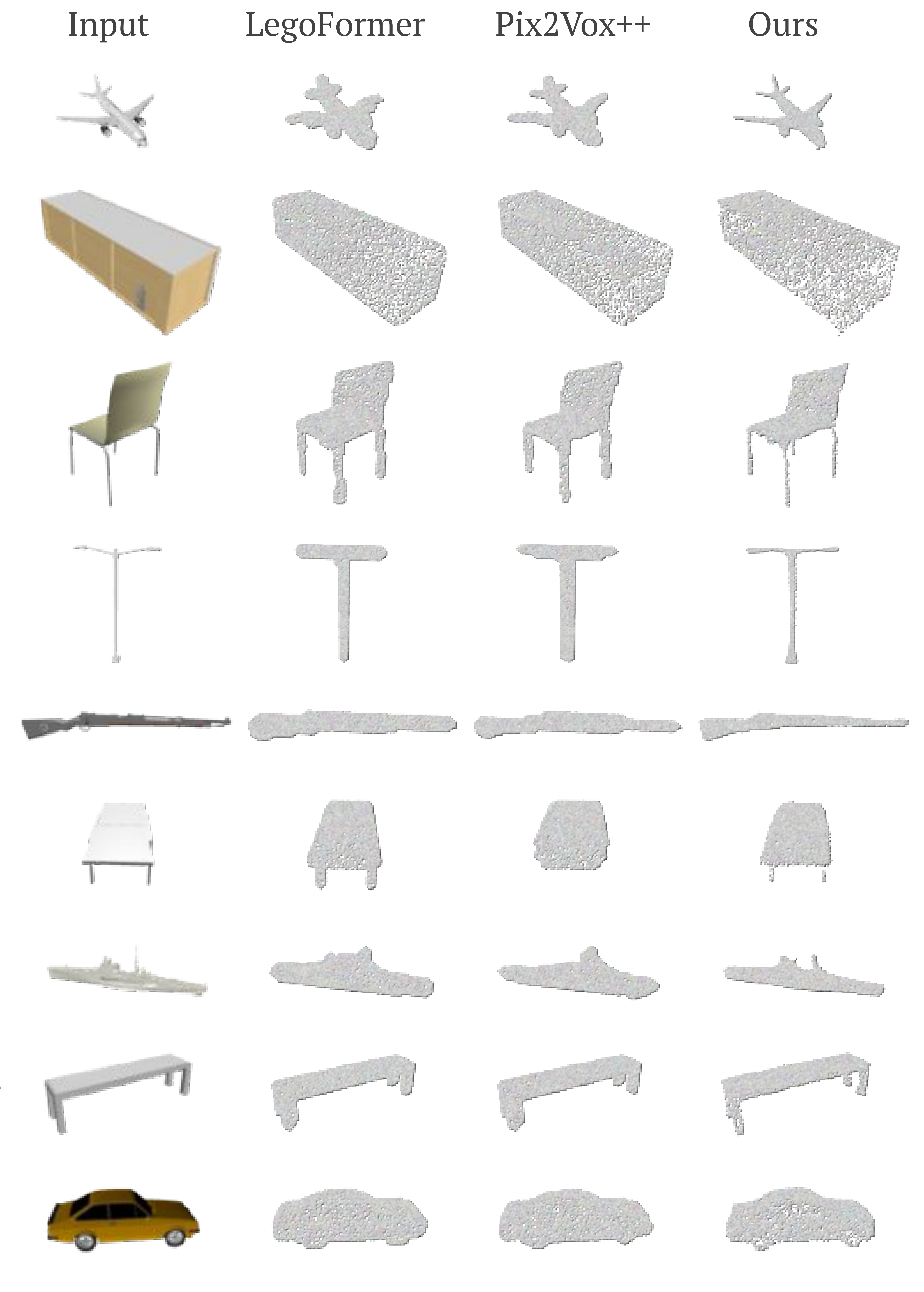}
\caption{Examples of reconstructions produced by our method along with prior work on seven classes from the ShapeNet-R2N2 dataset~\cite{choy20163d}.
The leftmost image in each row is the conditioning image.
Given the simplicity and synthetic nature of ShapeNet, all models produce adequate results;
however, our model produces reconstructions with the highest level of detail.
We perform particularly well on challenging categories such as airplane and rifle.
}%
\label{fig:comparison}
\end{figure}

\section{Experiments}\label{s:experiments}

We will first describe our datasets, model, and the implementation details.
We then discuss and analyze our results. 

\paragraph{ShapeNet.}

The ShapeNet Dataset is a collection of 3D CAD models corresponding to categories in the WordNet lexical database.
Due to its synthetic nature, ShapeNet is the standard dataset for quantitative evaluation of single-view reconstruction methods.
We use the subset of 13 ShapeNet categories from 3D-R2N2~\cite{choy20163d} along with their standard set of camera viewpoints, renderings, and train-test splits, which we collectively refer to as the ShapeNet-R2N2 benchmark.
For evaluation, we use the widely-adopted F-score metric proposed by Tatarchenko~\cite{Tatarchenko_2019_CVPR}.
See the supplementary material for further details on the ShapeNet and evaluation.

\paragraph{Co3D.}

The Co3D dataset is a challenging dataset consisting of multi-view images of real-world objects from common object categories.
For training data, we use point clouds generated from running COLMAP~\cite{schoenberger2016sfm,schoenberger2016mvs} on each image sequence.
Through these experiment, we demonstrate that our method is able to reconstruct \textit{real-world} objects using training data that is \textit{derived only from multi-view images processed with COLMAP}. 

We show results for three categories: hydrants, teddy bears, and toy trucks.
Since there is no prior work on single-view reconstruction of Co3D, we train a NeRF-WCE~\cite{Henzler_2021_CVPR} as a baseline and present qualitative results (\cref{fig:qualitative_nerf}).

\paragraph{Implementation Details.}

For our diffusion model $s_{\theta}$, we adopt a Point-Voxel CNN (PVCNN)~\cite{liu2019pvcnn} based on its success as described in \cite{Zhou_2021_ICCV}. This model processes a point cloud $X_t$ using two branches simultaneously: a point-based branch which processes points individually and a voxel-based branch which aggregates nearby points in a hierarchical manner. In this way, the network is able to understand both the global and local structure of a point cloud.

We implement our model in PyTorch\cite{paszke2019pytorch} and use the PyTorch3D library\cite{ravi2020accelerating} for rasterization during the projection conditioning phase. Our point-voxel diffusion model is trained with batch size 16 for a total of 100,000 steps. For feature extraction, we use MAE~\cite{he21masked}. For optimization, we use AdamW~\cite{kingma2015adam} with a $\beta = (0.9, 0.999)$ and a learning rate which is decayed linearly from $0.0002$ to $0$ over the course of training. 
We use images of size $137 \times 137$px and point clouds with $8192$ points, because prior work computes the F-Score metric using 8192 points. 
For our diffusion noise schedule, we use a linear schedule with warmup in which beta increases from $1 \cdot 10^{-5}$ to $8 \cdot 10^{-3}$. 
All experiments are performed on a single GPU.

\paragraph{Quantitative Results.}

In \cref{tab:comparison_others}, we show results for the 13 classes in the widely-used ShapeNet-R2N2 benchmark. 
Without filtering, \methodname performs on-par with prior work. Examining the performance across categories, we see that our method performs better on categories with objects that have fine details, such as rifle and airplane. 

With filtering, which is only possible due to the probabilistic nature of our conditional diffusion process, \methodnameplus consistently outperforms prior work; 
it improves upon the state-of-the-art on the majority of the object categories and on average. 

\paragraph{Qualitative Results.}

\Cref{fig:qualitative} and \cref{fig:qualitative_nerf} show qualitative results of our method on the real-world Co3D dataset.
\Cref{fig:qualitative} visually demonstrates the intermediate steps of the diffusion process as well as the final coloring step. \Cref{fig:qualitative_nerf} compares our generations to those from NeRF-WCE~\cite{Henzler_2021_CVPR}.
Since NeRF-WCE is deterministic, it struggles to model highly uncertain regions, and produces blurry images for novel views which are far from the reference view.
By contrast, our method produces realistic object shapes that can be viewed from any viewpoint. 
Relative to prior methods, our method is able to generate shapes with a significantly finer level of detail.

Many additional qualitative results are shown in the supplementary material.

\paragraph{Diversity of Generations.}

One key benefit of our probabilistic approach to 3D reconstruction is that we are able to generate multiple plausible shapes for a given reference view, as shown in \cref{fig:diversity} on the ShapeNet benchmark. This figure contains eight generations with different random seeds for three highly-ambiguous images in the chair category. Our method produces meaningful variations in shape, while always remaining consistent with the input image from the given view. This ability to sample makes it possible for us to develop our filtering method, which we analyze below.

\paragraph{Filtering Analysis.}

In \cref{tab:comparison_oracle}, we compare the performance of our two filtering methods against our single-sample performance and an oracle.
The oracle chooses the best sample for each input image according to its $F$-score; it provides an upper-bound to the improvements attainable via filtering. 
Our mask-free filtering method (\methodnameplusminus) improves results substantially and using masks (\methodnameplus) improves them further. However, gap between our method and the oracle suggests that there remains room for further improvement; this represents an interesting potential avenue for future work. 

In \cref{tab:comparison_filtering}, we compare our performance when filtering with different numbers of images.
We find that using just two images significantly improves results, and adding additional images leads to further performance gains. Naturally, more and more samples further increase the performance but with diminishing returns. 

\paragraph{Limitations.}

The primary limitation of our model is the need for point cloud ground truth for training. As mentioned in the introduction, this data is more accessible than commonly imagined, as it can be extracted from multi-view images or videos. However, point clouds obtained from these methods can be noisy; in Co3D, for example, many of the ground-truth point clouds for the hydrant category contain holes in locations that were not observed in the original video sequence. We find that our method is relatively robust to this type of noisy data, but we feel it is still important to discuss this limitation.

\section{Conclusions}\label{s:conclusions}

In this paper, we have proposed \methodname, a novel diffusion-based method for single-view 3D shape reconstruction. 
Our method iteratively reconstructs a shape by projecting image features onto a partially-denoised point cloud during the diffusion process. 
Empirically, our experiments demonstrate the effectiveness of \methodname at both reconstructing geometry and point color.
Quantitatively, we outperform prior methods on synthetic benchmarks. Qualitatively, we are able to reconstruct objects with high levels of detail from challenging real-world images.
With regard to future work, it would be interesting to scale our method to larger datasets and models.
Given the success of scaling diffusion models of 2D images (\eg, DALLE-2~\cite{ramesh2022hierarchical}), we hope this work is a step along the path to developing similar models for 3D reconstruction. 

\paragraph*{Ethics.}

We use the ShapeNet and CO3D datasets in a manner compatible with their terms.
The images do not contain personal data.
ShapeNet models are used in a manner compatible with the Data Analysis Permission.
Please see \url{https://www.robots.ox.ac.uk/~vedaldi/research/union/ethics.html} for further information on ethics.

\paragraph*{Acknowledgments.}

L\@. M\@. K\@. is supported by the Rhodes Trust.
A\@. V\@. and C\@. R\@. are supported by ERC-UNION-CoG-101001212.
C\@. R\@. is also supported by VisualAI EP/T028572/1.

{\small\bibliographystyle{ieee_fullname}\bibliography{refs,refs_diffusion,vedaldi_general,vedaldi_specific}}

\clearpage
\section{Implementation Details}\label{s:supplement_implementation}

Here, we provide additional details about the model and projection procedure. 

First, we discuss the Point-Voxel~\cite{liu2019pvcnn} model which is used to process the partially-denoised point cloud at each step of the diffusion process. As its name suggests, this model processes a point cloud using two branches simultaneously: a point-based branch and voxel-based branch. The point-based branch is a simple multi-layer perceptron which is applied to each point independently, as in PointNet~\cite{qi2017pointnet,qi2017pointnetpp} (without the global pooling in the final layer of PointNet). The voxel-based branch first discretizes the points into a coarse voxel grid of size $128^3$, which is fed into a 3D U-Net. As in \cite{liu2019pvcnn}, the 3D U-Net consists of four downsampling (``Set Abstraction'') layers followed by four upsampling (``Feature Propogation'') layers. Due to this fine-to-coarse-to-fine structure, the network is able to capture both global and local shape information. Additionally, to make the model aware of the current timestep of the diffusion process, we concatenate an embedding of the current timestep to the point features at the input to each layer.

Second, we discuss the implementation of the projection feature. We perform the projection by rasterizing the point cloud from the given camera view. We utilize the \texttt{PointRasterizer} class of PyTorch3D using a point radius of $0.0075$ and $1$ point per pixel. For each point in the point cloud, if the point is rasterized onto a pixel in the input image, we concatenate the image features corresponding to the pixel onto that point's existing feature vector (which is simply a sinusoidal positional embedding of its current position) for input to the model. Additionally, we concatenate the value of the (binary) object mask at the given pixel and a two-dimensional vector pointing from the pixel to the closest pixel in the mask (i.e. a two-dimensional distance function corresponding to the mask region; this is the zero vector inside the mask and a nonzero vector outside the mask). If a pixel is not rasterized to a point (for example, because it is occluded by another point), we concatenate a vector of zeros in place of all the quantities above. 

\section{Additional Qualitative Examples}\label{s:supplement_qualitative}

\begin{figure*}
  \centering
  \includegraphics[width=\textwidth]{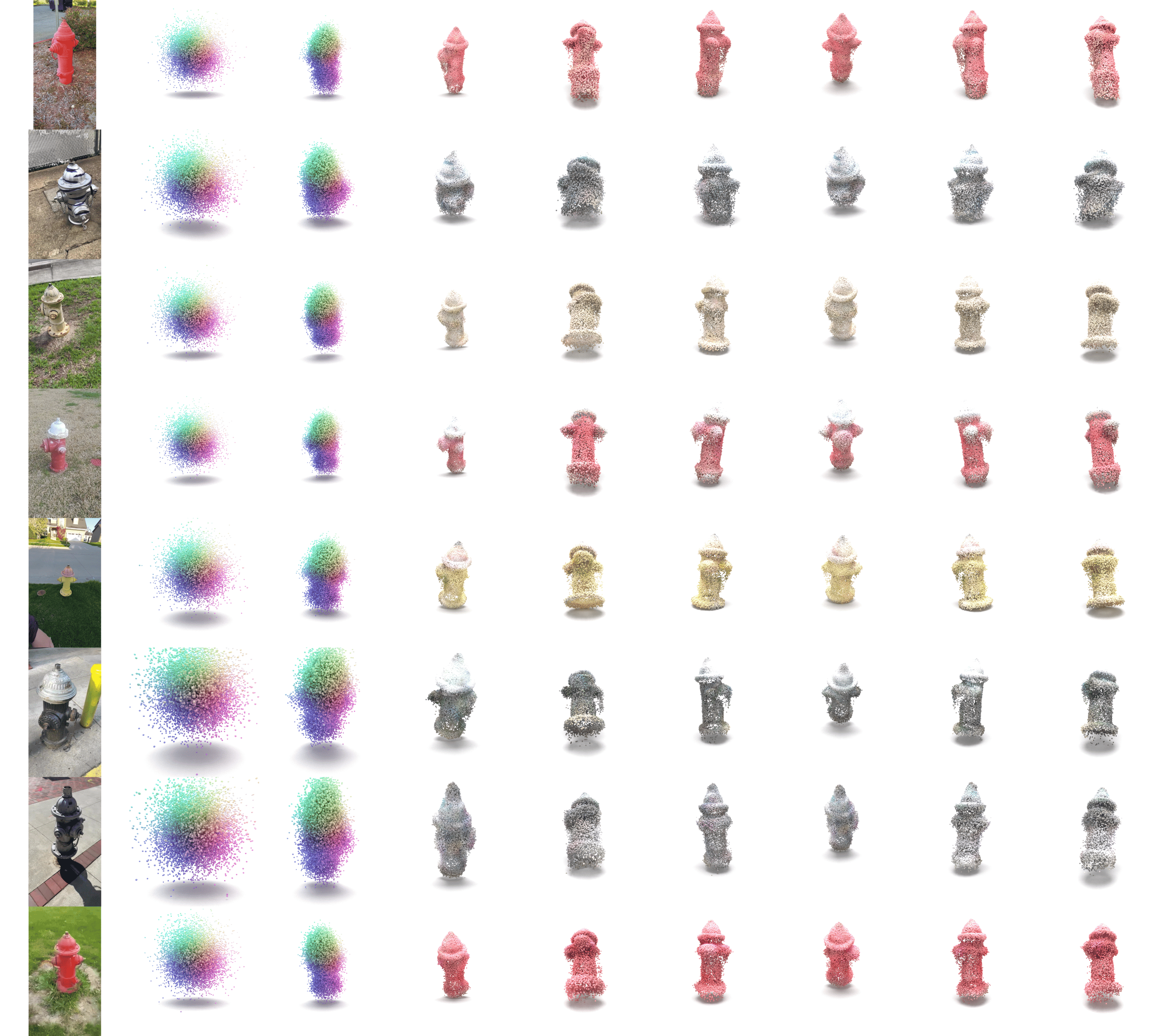} \\ \vspace{7mm}
  \caption{
  \textbf{Additional qualitative examples.} 
  Examples from the hydrants category of Co3D. The first column in each row shows the input image. The second and third columns show intermediate steps in the diffusion process. The fourth column shows the final reconstructed point cloud with color. The remaining five rows show the final predicted point cloud from novel views.
  }
  \label{fig:supp_examples_hydrant}
\end{figure*}

\begin{figure*}
  \centering
  \includegraphics[width=\textwidth]{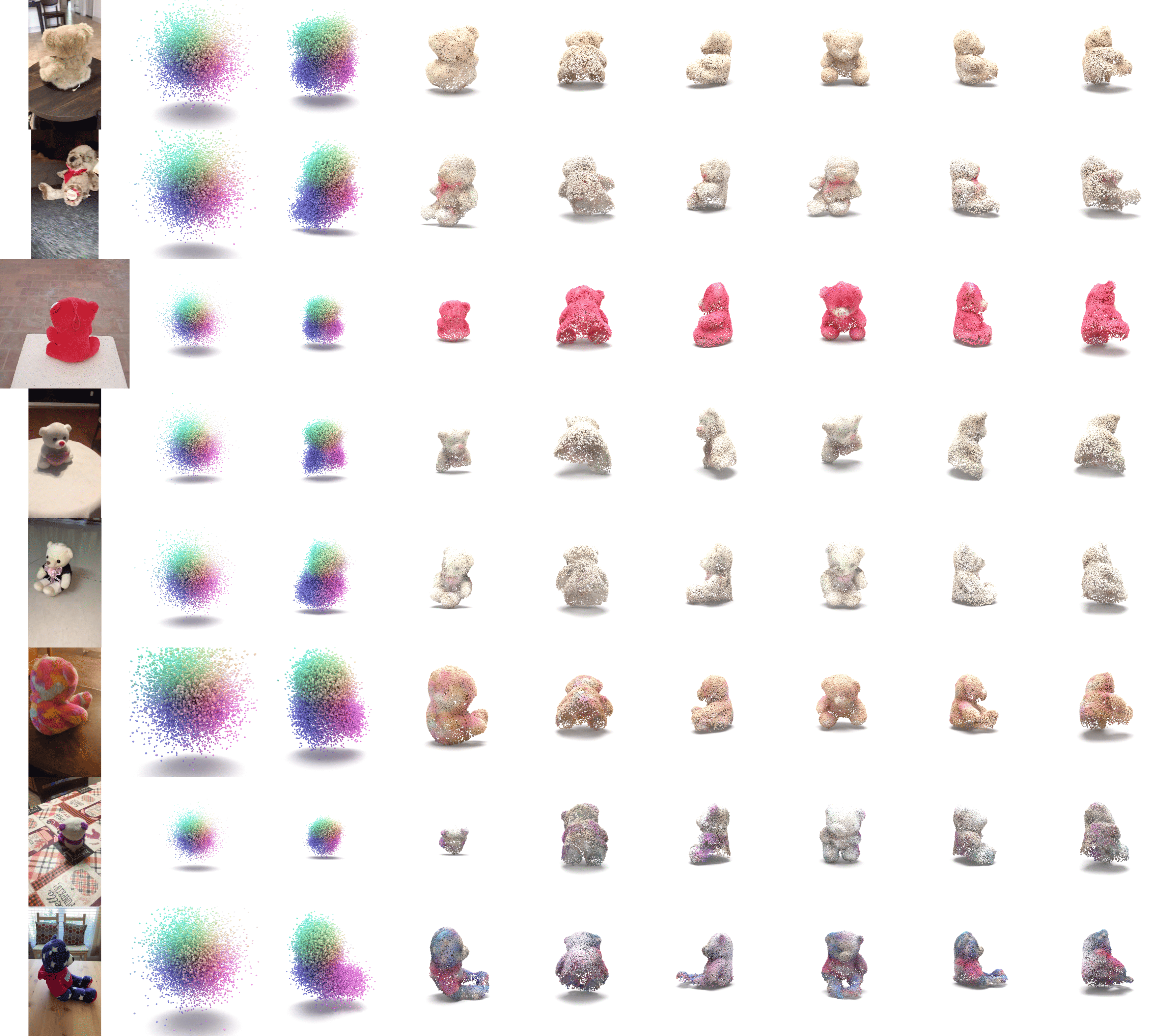} \\ \vspace{7mm}
  \caption{
  \textbf{Additional qualitative examples.} 
  Examples from the teddy bear category of Co3D. The first column in each row shows the input image. The second and third columns show intermediate steps in the diffusion process. The fourth column shows the final reconstructed point cloud with color. The remaining five rows show the final predicted point cloud from novel views.
  }
  \label{fig:supp_examples_teddybear}
\end{figure*}

\begin{figure*}
  \centering
  \includegraphics[width=0.8\textwidth]{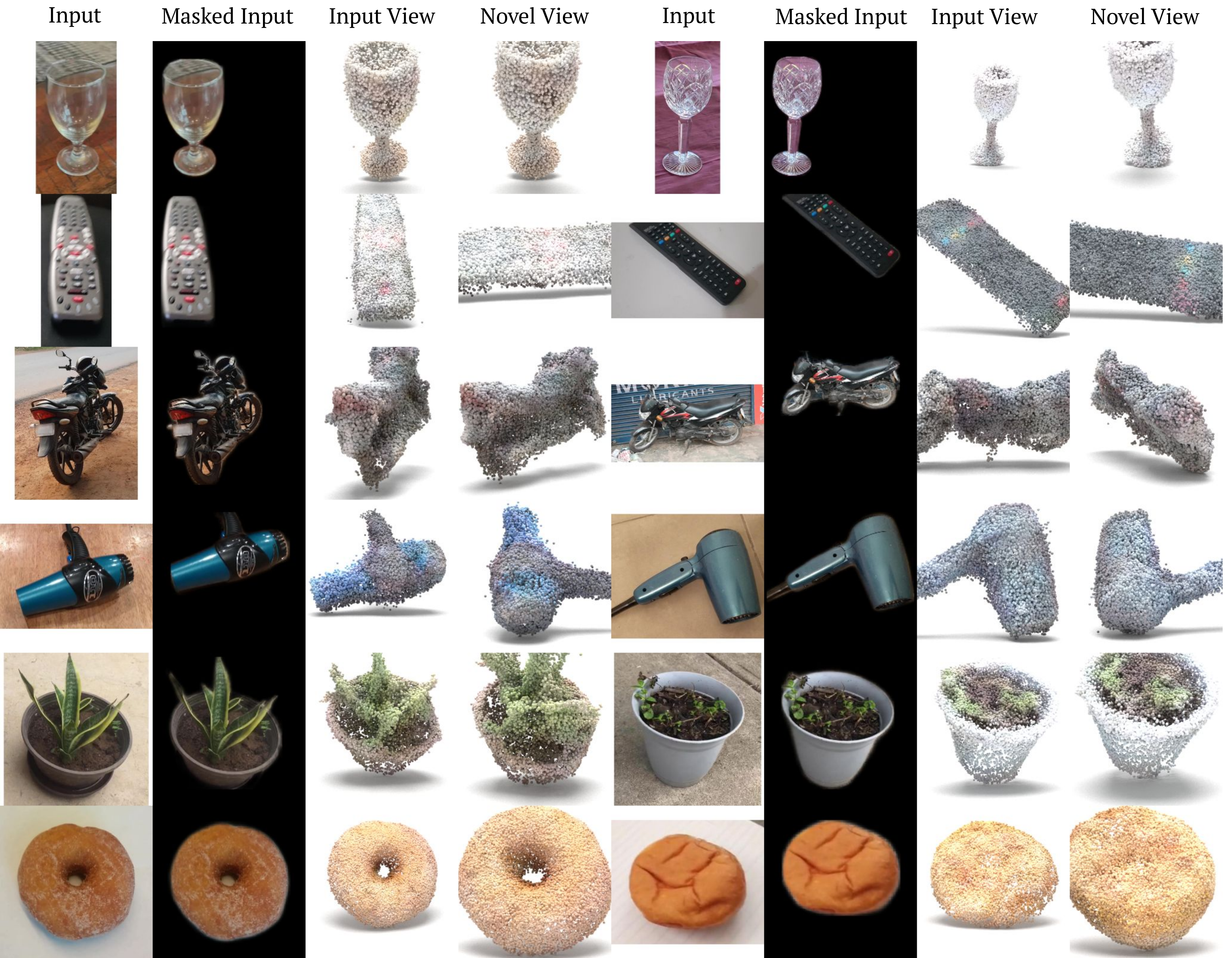} \\ \vspace{7mm}
  \caption{
    \textbf{Additional qualitative examples.} 
    Examples from six additional categories: glasses, remotes, motorcycles, hairdryers, plants, and donuts.
  }
  \label{fig:supp_examples_more}
\end{figure*}

\begin{figure*}
  \centering
  \includegraphics[width=\textwidth]{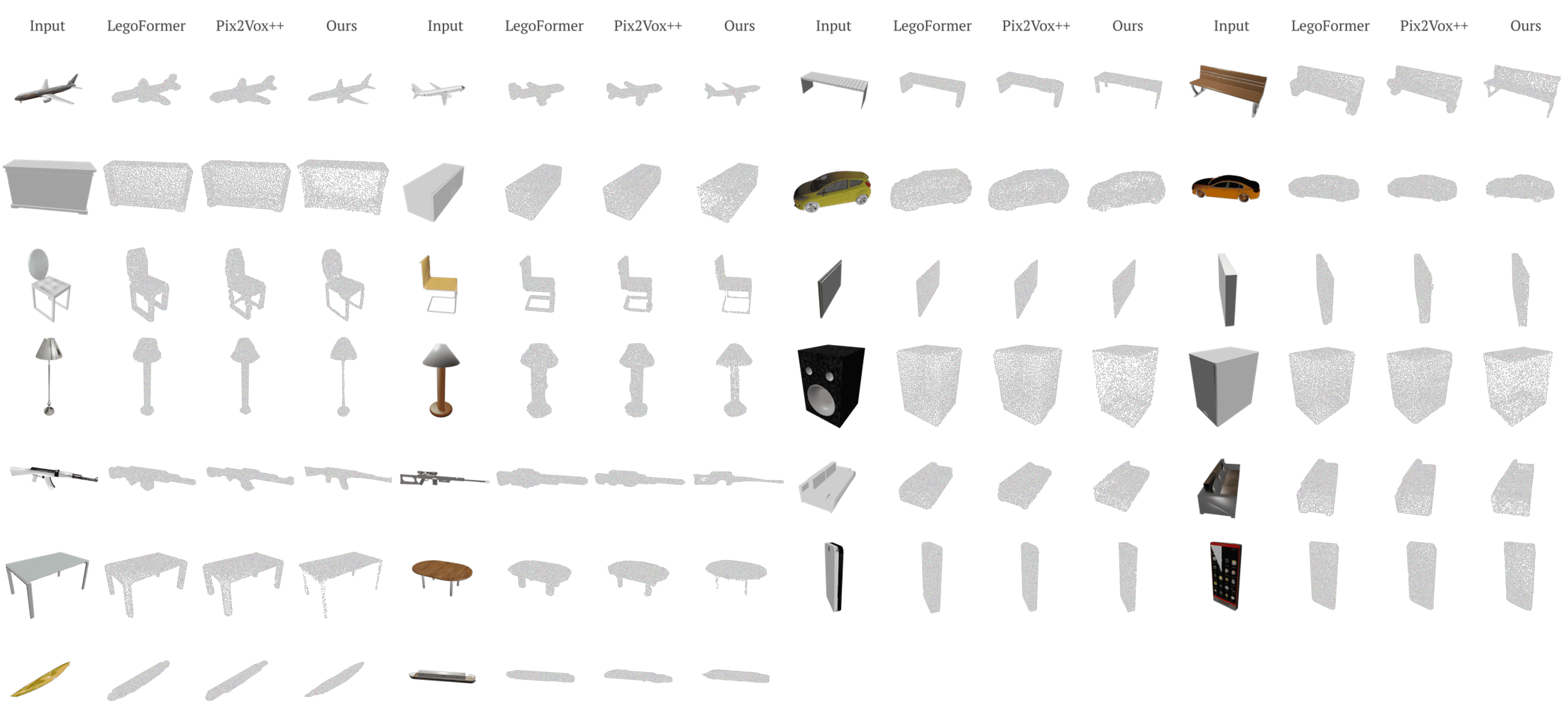} \\ \vspace{7mm}
  \includegraphics[width=\textwidth]{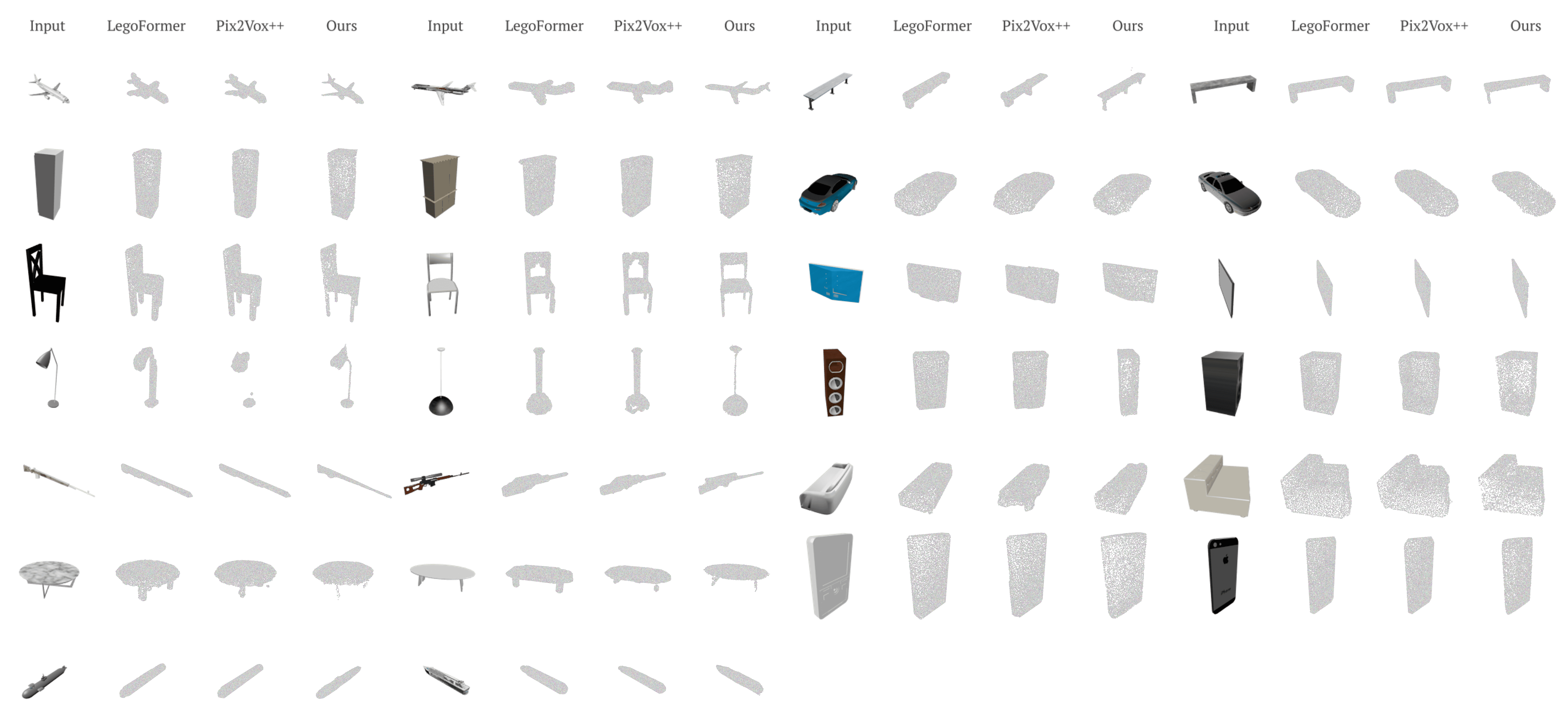} \\ \vspace{7mm}
  \caption{\textbf{Successful examples} produced by our method along with prior work. The leftmost image in each set of images is the input image. Note that there are no images in the last row of the right half of the figure because we show examples for all 13 ShapeNet-R2N2 classes (seven on the left and six on the right).
  }\label{fig:supp_examples_best_1}
\end{figure*}

\begin{figure*}
  \centering
  \includegraphics[width=\textwidth]{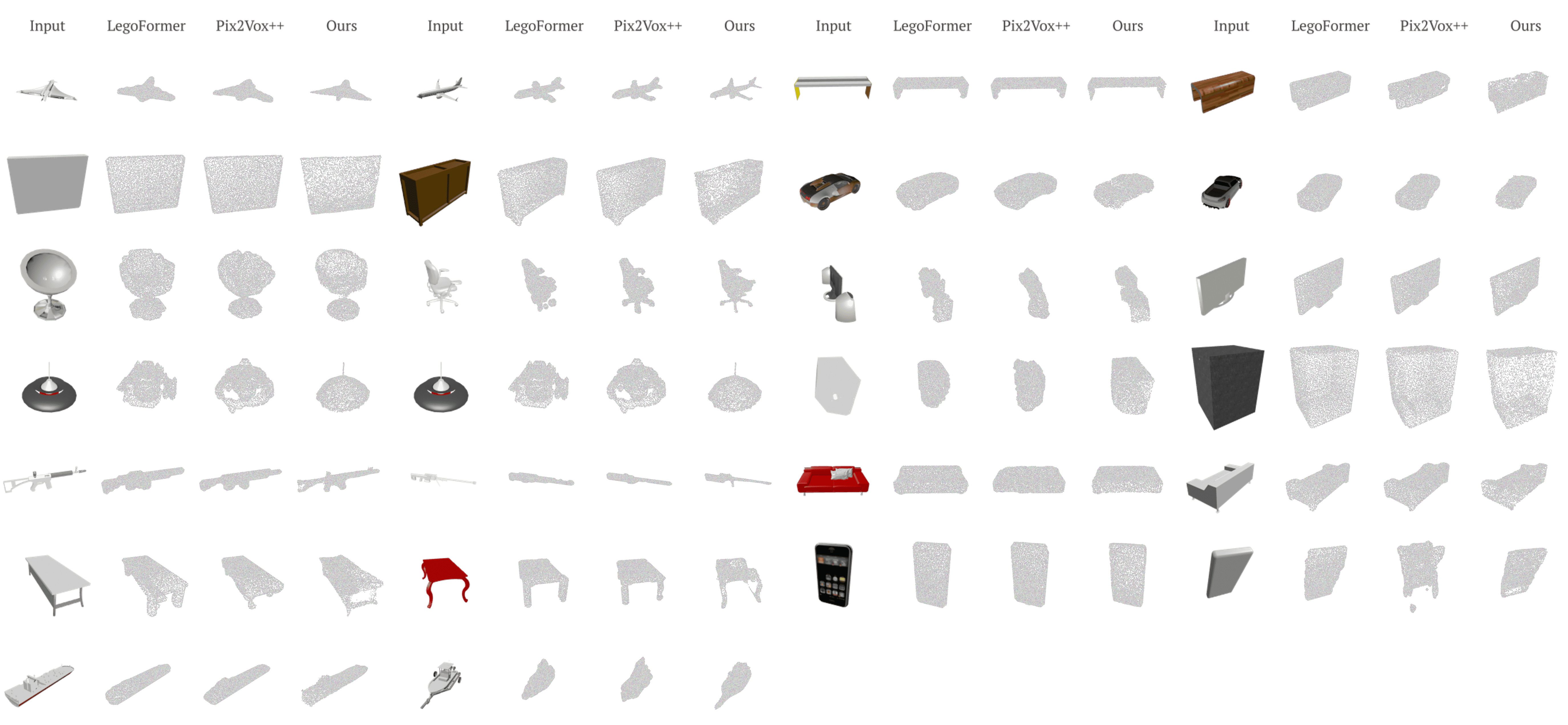} \\ \vspace{7mm}
  \includegraphics[width=\textwidth]{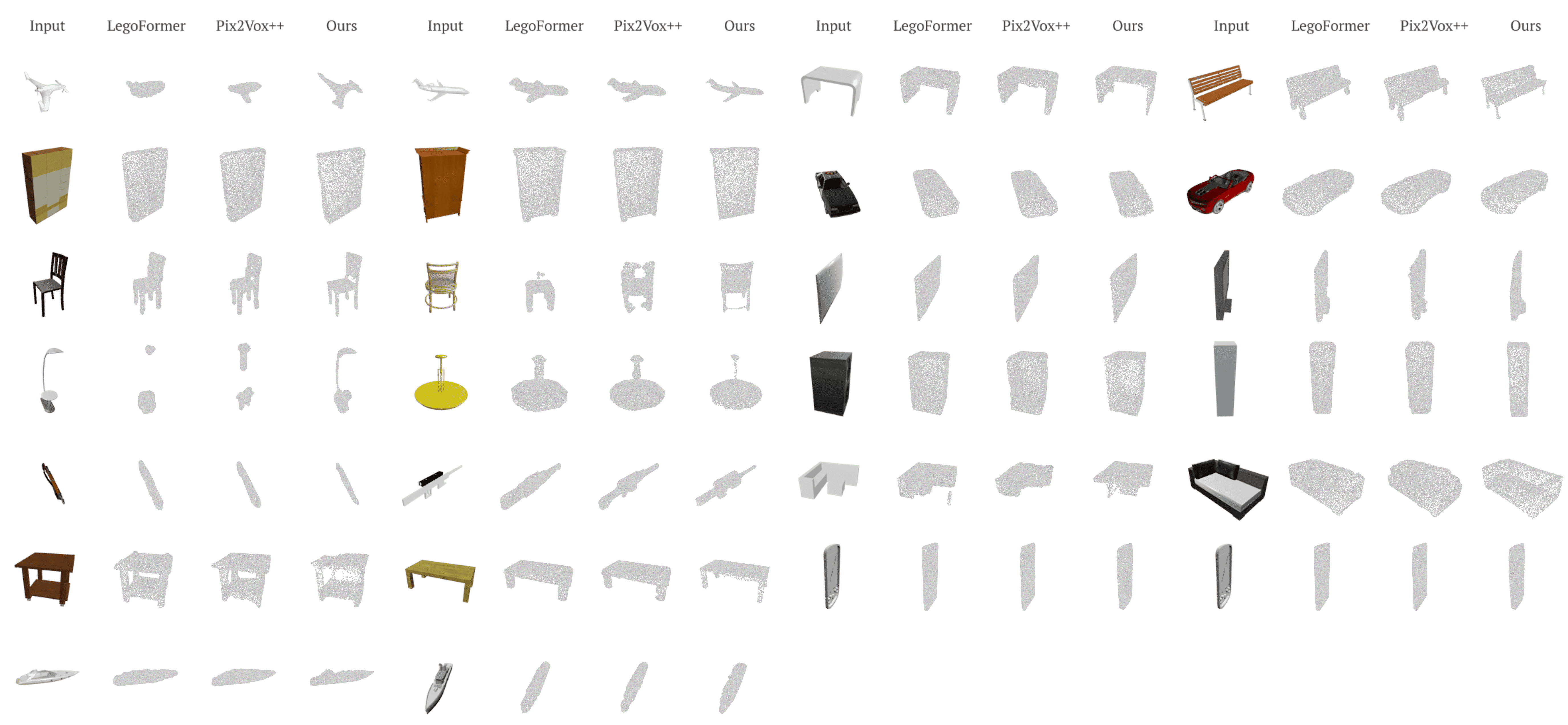} \\ \vspace{7mm}
  \caption{\textbf{Random examples} of our method along with prior work. The leftmost image in each set of images is the input image. Note that there are no images in the last row of the right half of the figure because we show examples for all 13 ShapeNet-R2N2 classes (seven on the left and six on the right).}\label{fig:supp_examples_random_1}
\end{figure*}

We provide additional qualitative examples of our method in 
\cref{fig:supp_examples_hydrant,fig:supp_examples_teddybear,fig:supp_examples_best_1,fig:supp_examples_worst_2,fig:supp_examples_random_1}.
\Cref{fig:supp_examples_hydrant,fig:supp_examples_teddybear,fig:supp_examples_more} show examples of reconstructions on additional categories of Co3D, including hydrants, teddybears, glasses, remotes, motorcycles, hairdryers, plants, and donuts.
\cref{fig:supp_examples_best_1} contains a selection of the best reconstructions produced by our model for each category of ShapeNet, as ranked by F-score. \cref{fig:supp_examples_random_1} contains random examples of reconstructions produced by our model on ShapeNet. Finally, \cref{fig:supp_examples_worst_2} shows a selection of the worst examples produced by our model for each category on ShapeNet, as ranked by F-score. 

\section{Additional Ablations}\label{s:supplement_ablations}

We include additional ablations omitted from the main paper due to space constraints. These ablations were performed on a subset of ShapeNet dataset consisting of only the \texttt{sofa} category.

\paragraph{Mask Distance Function.} We removed the 2D mask distance function described in \Cref{s:supplement_implementation}. This change had a small effect, reducing the $F$-score by $0.019$ points, a relative decrease of 9\%. Qualitatively, the generated point clouds were similar to those produced using the mask distance function.

\paragraph{Projection Method.} We replaced the rasterization-based projection described in Section 3.4 with a naive projection that projects all points (including occluded points) onto the image. This change was detrimental, reducing the $F$-score by $0.081$ points, a relative decrease of 40\%. Qualitatively, these point clouds were significantly worse than those with the rasterization-based projection. These results suggest that the rasterization-based projection is a key component of the method.  

\section{Analysis of Failure Cases}\label{s:supplement_failure_ases}

\begin{figure*}
  \centering
  \includegraphics[width=\textwidth]{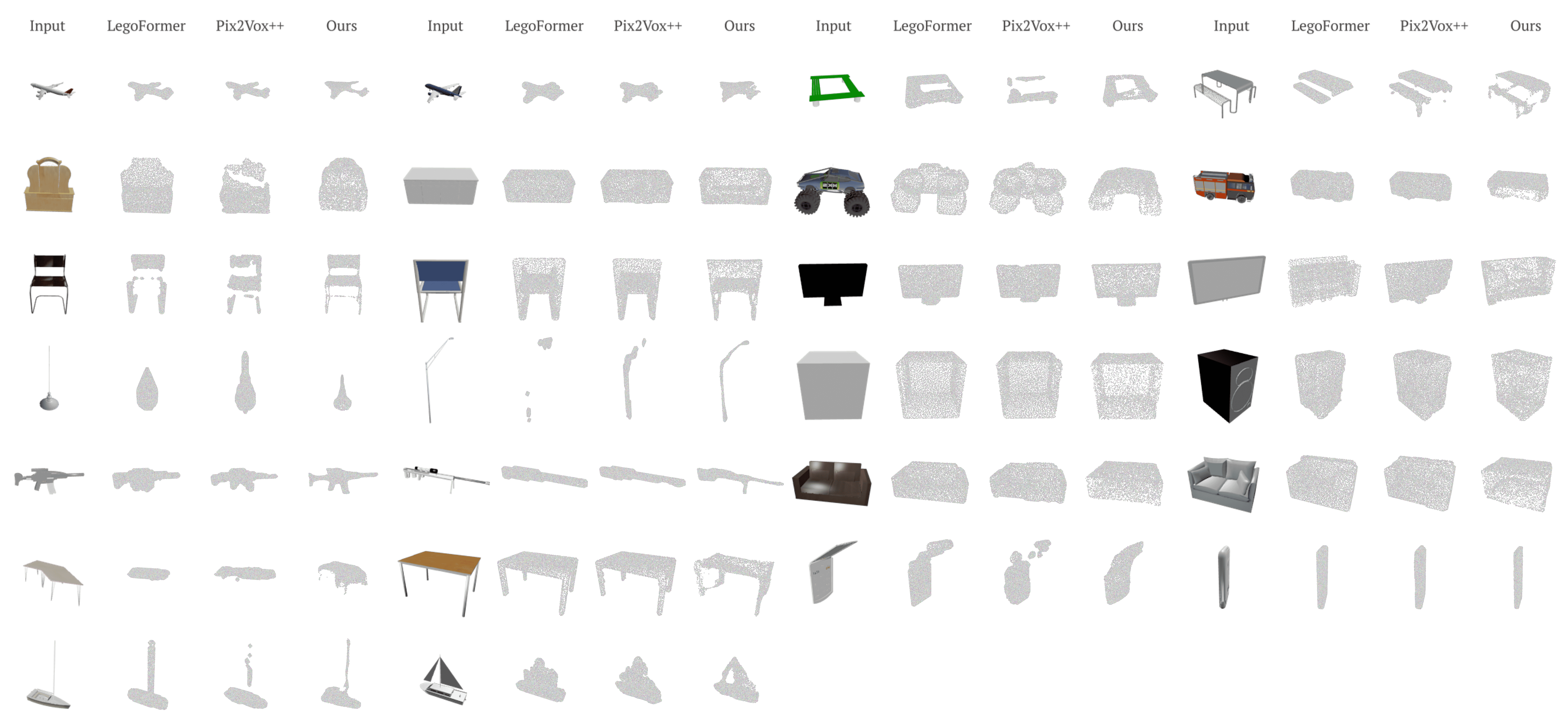} \\ \vspace{7mm}
  \includegraphics[width=\textwidth]{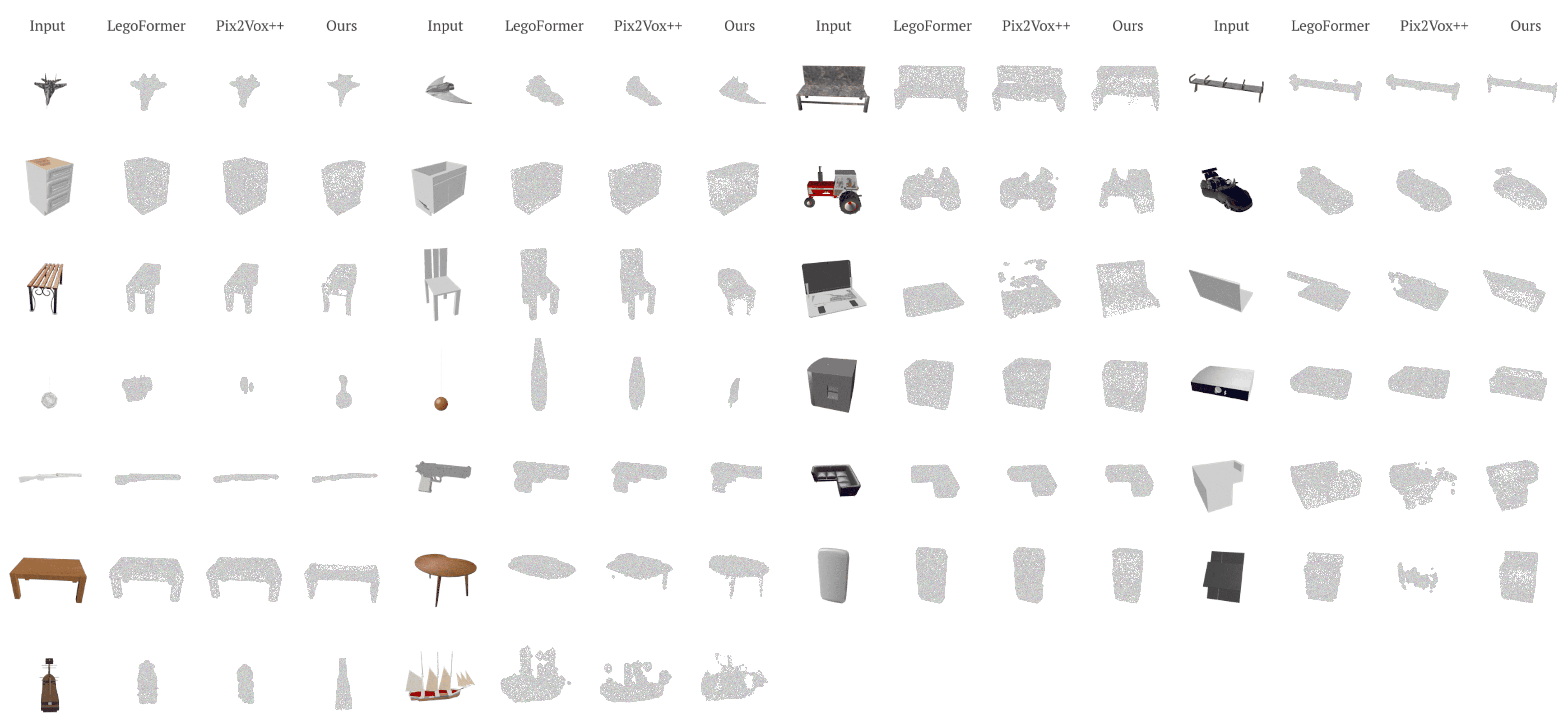} \\ \vspace{7mm}
  \caption{\textbf{Failure cases} of our method along with prior work. The leftmost image in each set of images is the input image. Note that there are no images in the last row of the right half of the figure because we show examples for all 13 ShapeNet-R2N2 classes (seven on the left and six on the right).}\label{fig:supp_examples_worst_2}
\end{figure*}

Failure cases of our model are shown in \cref{fig:supp_examples_worst_2}. Note that these are from the ShapeNet-R2N2 dataset, which combines 13 ShapeNet classes but \textit{does not} permit the use of category labels. In other words, the model is image-conditional, but not class-conditional. 

Examining these failure cases, we observe that our model sometimes performs poorly on images with ambiguous categories. For example, in the $8^{th}$ column of the $2^{nd}$ row of the figure, it appears that the model generates a chair rather than a box. Similarly, in the $12^{th}$ row of the $5^{th}$ row of the figure, the object seems to have generated a box rather than a couch. These errors are most likely due to the fact that these categories all have instances which resemble rectangular prisms from certain views. 

It is also notable that on many of the challenging examples on which our model struggles (\eg, the examples for the \texttt{watercraft} category located in the last row of the figure), other models also struggle to a similar degree.

\end{document}